\documentclass[journal,twoside,web]{ieeecolor}
\usepackage{generic}

\usepackage{multirow}
\usepackage{amsmath}
\usepackage{amssymb}
\usepackage{amsfonts}
\usepackage{algorithm}
\usepackage{algorithmic}
\usepackage{booktabs}
\usepackage[table]{xcolor}
\usepackage{tabularx}

\makeatletter
\if@web
  \definecolor{nblue}{rgb}{0,0.263,0.576}
  \definecolor{mblue}{rgb}{0.075,0.541,0.855}
\else
  \definecolor{nblue}{cmyk}{0,0,0,1}
  \definecolor{mblue}{cmyk}{0,0,0,1}
\fi
\makeatother
\definecolor{subsectioncolor}{rgb}{0,0,0}
\usepackage{graphicx}

\makeatletter
\let\NAT@parse\undefined
\makeatother
\usepackage{hyperref}

\hypersetup{
    colorlinks=true,
    citecolor=blue,
    linkcolor=blue,
    urlcolor=blue,
    filecolor=black,
    breaklinks=true
}
\usepackage{textcomp}

\providecommand{\refname}{REFERENCES}

\usepackage{etoolbox}
\patchcmd{\thebibliography}{0pt plus pt}{0pt plus 2pt}
  {}{\typeout{AFOR warning: bibliography itemsep patch did not apply}}

\patchcmd{\thebibliography}{\section*{\color{black}}}{\section*{\color{black}\refname}}
  {}{\typeout{AFOR warning: bibliography heading patch did not apply}}

\title{Adaptive Forgetting for Nonstationary Optimization: Towards Robust EEG Decoding}

\author{Hongyu Zhu, Lin Chen, Jing Chen, Yuting Zhou, and Mingsheng Shang%
\thanks{H. Zhu, L. Chen, and M. Shang are with Chongqing Institute of Green Intelligent Technology, Chinese Academy of Sciences, e-mail: \{zhuhongyu, chenlin, msshang\}@cigit.ac.cn. (Corresponding authors: Lin Chen; Mingsheng Shang).}%
\thanks{Y. Zhou is with School of Resources and Environment, University of Electronic Science and Technology of China, e-mail: ytzhou@std.uestc.edu.cn.}%
\thanks{J. Chen is with Beijing University of Posts and Telecommunications, e-mail: jcccchen@163.com.}
\thanks{This work was supported in part by the Chongqing Key Project of Technological Innovation and Application (CSTB2023TIAD-STX0015, CSTB2025TIAD-STX0034) and the National Natural Science Foundation of China (Grant No. 62506054).}%
\thanks{CODE: \url{https://github.com/HongyuZhu-s/AFOR}.}%
}

\begin{document}

\maketitle

\begin{abstract}
Electroencephalography (EEG) provides non-invasive monitoring of brain activity and is widely used in emotion recognition, motor imagery and sleep staging.
Although within-subject decoding has achieved considerable progress, cross-subject generalization remains a central challenge in practical applications. EEG decoders are typically trained with Adam/AdamW under a fixed second-moment decay coefficient, even though cross-subject learning involves low signal-to-noise ratios, subject variability, and gradient nonstationarity. A fixed coefficient implicitly assumes that gradient statistics are homogeneous across layers and time, which can limit model's adaptability to cross-subject EEG signals and degrade generalization.
To address these issues, we propose AFOR, a tensor-wise adaptive optimizer that converts the fixed second-moment decay coefficient into a dynamic coefficient estimated online from local gradient state. AFOR combines a Residual-Alignment Signal Scorer (RASS) and an Adaptive Forgetting Controller (AFC). RASS summarizes local gradient residuals and directional agreement into a signal-quality score, and AFC maps this score through self-referential normalization to a bounded per-step decay coefficient, with cumulative-product initialization correction maintaining consistency under time-varying decay.

Under a strict cross-subject protocol on three EEG benchmarks that cover three representative fields, AFOR achieves the best average performance among the compared optimizers, improving the mean test
accuracy over Adam by 3.00\%, 2.07\%, and 4.38\%, respectively.
\end{abstract}

\begin{IEEEkeywords}
EEG, cross-subject, adaptive optimizer.
\end{IEEEkeywords}

\section{INTRODUCTION}

\IEEEPARstart{E}{lectroencephalography} (EEG) is a noninvasive technique that records brain activity with millisecond resolution, making it an important modality in brain-computer interfaces (BCIs) and cognitive neuroscience~\cite{marino2026human}. It is widely used for tasks such as emotion recognition, motor imagery, and sleep staging. Deep learning models, including Convolutional Neural Networks (CNNs)~\cite{eegnet, 2022TSception}, Recurrent Neural Networks (RNNs)~\cite{acrnn, 2023EEGRNN2}, Graph Neural Networks (GNNs)~\cite{nsaldgat, EEGGNN2}, and transformer~\cite{hslt, eegtrans2} architectures, have become the standard approach to decoding EEG signals into behavioral states. This lays the foundation for more intelligent and reliable human-computer interaction (HCI)~\cite{EEGBCI2025}.

Although within-subject decoding has achieved substantial progress~\cite{yang2026dual,yang2026prototypical}, EEG decoding models still often struggle to generalize to unseen subjects, which limits their deployment in real-world scenarios. Improving cross-subject generalization is therefore essential~\cite{li2026cross}. Compared with modifying model architectures or the collection of additional labeled EEG data, optimizing the training procedure can provide a comparatively low-cost way to improve cross-subject generalization~\cite{ng2024subject}. However, most EEG-based works~\cite{libeer} still default to Adam~\cite{adam} or AdamW~\cite{adamw}, both of which use a fixed second-moment decay coefficient. This fixed-memory design assumes that gradient statistics are stable enough across layers and training stages to share one global memory length. Yet in EEG decoding, this assumption may not hold because high inter-subject signal variability~\cite{ng2024subject} and annotation uncertainty~\cite{zhu2026phy} make gradient statistics highly heterogeneous across subjects and training stages. This heterogeneity can lead to nonstationary gradient signals that undermine the fixed-memory assumption and ultimately limit cross-subject generalization.

Existing optimizer work can be grouped into two categories. The first improves the update rule itself, such as stabilizing early steps or reducing variance~\cite{radam,adamp,sophia,mars}, but it still keeps the second-moment memory fixed, so it cannot adapt the second-moment decay coefficient to local gradient changes. The second studies variable forgetting in nonstationary settings~\cite{paleologu2008robust,huang2019nostalgic}, but its decay schedules are still global or predetermined, making them insensitive to tensor-wise gradient state. Neither approach reacts when the gradient statistics change, so updates for a new subject are still scaled by statistics accumulated from earlier ones. The same memory length applies to the whole network, while layers respond to subject shift at different rates.

To address this limitation, we propose a tensor-wise \textbf{A}daptive \textbf{F}orgetting \textbf{O}ptimize\textbf{r} (AFOR) that replaces the fixed second-moment decay coefficient with a per-tensor, per-step value estimated online from the local gradient. In cross-subject training, mini-batches are drawn from a few subjects at a time, so the second-moment estimate lags the current subject; shortening the memory only trades that lag for noise. Because the tradeoff differs across tensors and steps, AFOR sets the coefficient from the gradient itself, using a \textbf{R}esidual-\textbf{A}lignment \textbf{S}ignal \textbf{S}corer (RASS) and an \textbf{A}daptive \textbf{F}orgetting \textbf{C}ontroller (AFC).

\begin{figure*}[!t]
\centering
\includegraphics[width=\textwidth]{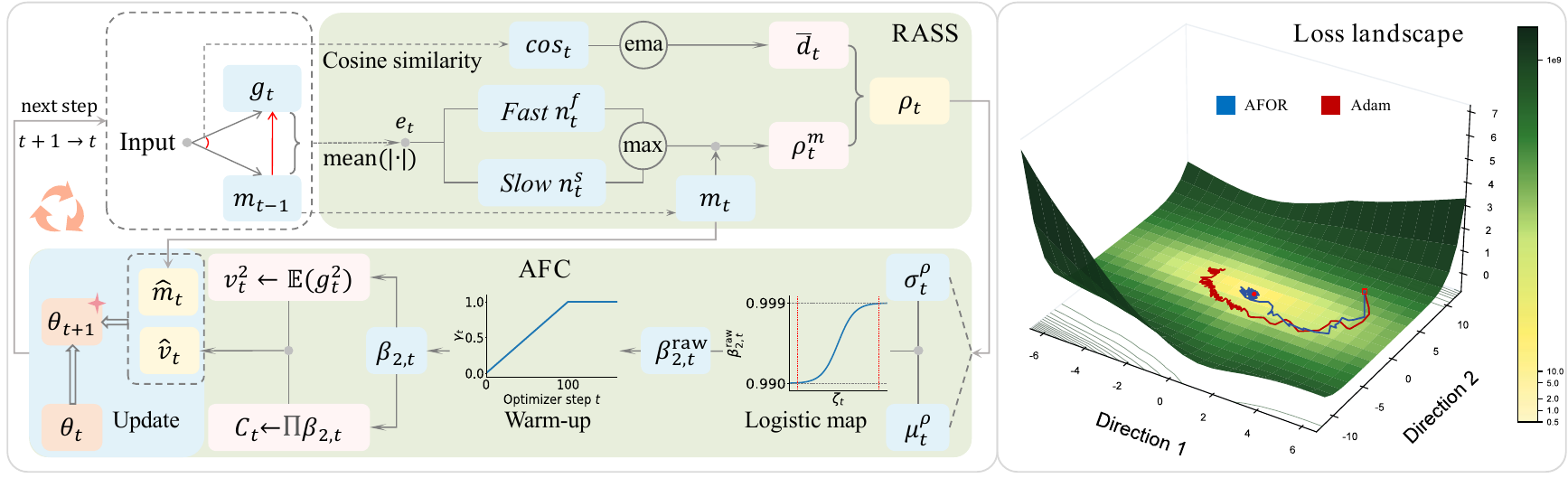}
\caption{Two-part overview of AFOR. The left panel shows the implementation flow of the optimizer, where local gradient statistics are converted into a tensor-wise, step-wise $\beta_{2,t}$. The right panel shows the loss landscape of EEGNet on ISRUC-Sleep SG-I, centered at the AFOR solution, where the red curve denotes Adam and the blue curve denotes AFOR.}
\label{fig:method}
\end{figure*}

Our contributions are as follows:

\begin{itemize}
\item We propose \textbf{AFOR}, a tensor-wise adaptive optimizer for cross-subject EEG decoding that replaces the fixed or globally scheduled second-moment rates of conventional adaptive optimizers with online, state-dependent memory regulation.
\item We develop a Residual-Alignment Signal Scorer (\textbf{RASS}), which provides a normalized, tensor-wise measure of update reliability by jointly tracking residual fluctuations and directional consistency.
\item We design an Adaptive Forgetting Controller (\textbf{AFC}), which enables
tensor-wise and step-wise regulation of the second-moment memory according to
local gradient quality, allowing AFOR to shorten memory for unreliable updates
and retain longer history for stable ones under nonstationary optimization.
\item We evaluate AFOR under a strict cross-subject protocol on three EEG benchmarks covering emotion recognition, motor imagery, and sleep staging. Across four representative backbones and six optimizers, AFOR achieves the best average performance.

\end{itemize}

\section{RELATED WORK}

\subsection{EEG Decoding and Benchmarking}

EEG decoding has evolved from handcrafted pipelines to deep learning pipelines, with models typically built upon CNNs, RNNs, GNNs, and Transformers~\cite{libeer}. These deep learning-based decoding models have been widely applied to emotion recognition~\cite{zhu2026bimoe,ding2025emt}, sleep staging~\cite{phang2025explainable,wang2026brastorm}, and motor imagery~\cite{saibene2024mireview,rao2024wearable}, among other tasks~\cite{kostulin2026eeg}. EEGNet~\cite{eegnet} established a lightweight CNN baseline for cross-subject BCIs, and subsequent works extended this direction with sparse and attention-augmented CNNs~\cite{rakhmatulin2024cnn,eegnetmsd2023}, domain-adaptive graph modeling~\cite{nsaldgat}, dynamic GNNs for seizure classification~\cite{hajisafi2024neurognn}, and hierarchical transformers~\cite{hslt}. In recent work, Yang et al.~\cite{yang2026dual} proposed a dual-stream regional feature learning and adaptive fusion network that integrates local and global EEG representations for emotion recognition. Benchmarks and libraries such as DEAP~\cite{2011deap}, SEED~\cite{seed}, LibEER~\cite{libeer}, BCI Competition IV 2a (BCI IV)~\cite{bci42a}, and ISRUC-Sleep SG-I (ISRUC)~\cite{isruc} have standardized preprocessing and evaluation across emotion, motor-imagery, and sleep-staging tasks. These studies indicate that while architecture choice matters, cross-subject generalization also depends heavily on how the model is optimized under subject variability. A fixed optimizer, however, cannot adapt its behavior to such variability, and may thus fall short in cross-subject settings.

\subsection{Adaptive Optimizers and Forgetting Mechanisms}

In EEG-based deep learning, Adam~\cite{adam} and AdamW~\cite{adamw} serve as the default optimizers. Prior work can be roughly divided into two directions. The first direction revises the adaptive update rule. RAdam~\cite{radam} stabilizes early training when variance estimates are noisy, AdamP~\cite{adamp} adjusts the update on scale-invariant weights, Sophia~\cite{sophia} replaces first-order variance estimates with a lightweight second-order proxy, and MARS~\cite{mars} reduces gradient variance through recursive correction. These methods improve specific aspects of optimization, yet they still rely on a fixed global second-moment memory. As a result, the second-moment decay coefficient remains unresponsive to local gradient conditions.
The second direction studies forgetting mechanisms for nonstationary optimization~\cite{paleologu2008robust,huang2019nostalgic}. They move closer to adaptive memory, but the schedule is typically global or predetermined, which prevents it from tracking tensor-wise changes in gradient noise. This leaves the cross-subject, layer-wise nonstationarity of EEG decoding inadequately addressed by existing optimizer designs.

\section{METHOD}

AFOR is a tensor-wise adaptive forgetting optimizer for EEG decoding. Its core idea is to replace a fixed second-moment decay coefficient with a dynamic tensor-wise coefficient estimated online from local gradient state. 
The optimizer consists of two coupled components. \textbf{RASS} summarizes the deviation between the current gradient and its momentum estimate, together with their directional consistency, into a scalar signal-quality score. \textbf{AFC} then maps this score through self-referential normalization to a bounded second-moment decay coefficient and uses cumulative-product tracking to preserve the initialization-correction factor under time-varying decay.
Together these components form the closed-loop design shown in Fig.~\ref{fig:method}, which adjusts memory length per tensor according to the observed gradient dynamics rather than relying on a global schedule.

\subsection{Problem Formulation}

Let $\mathcal{D} = \{\mathcal{D}_s\}_{s=1}^{S}$ denote an EEG dataset collected from $S$ subjects, where $\mathcal{D}_s = \{(X_{s,i}, y_{s,i})\}_{i=1}^{N_s}$ is the set of $N_s$ samples from subject $s$. Each sample consists of an input $X_{s,i} \in \mathbb{R}^{C \times T_x}$ with $C$ channels and $T_x$ time points and a label $y_{s,i} \in \{1,\dots,K\}$. Our goal is to train a neural classifier $f_\theta: \mathbb{R}^{C \times T_x} \to \mathbb{R}^K$ with parameters $\theta \in \mathbb{R}^d$, which maps each input $X_{s,i}$ to a logit vector. For a given cross-subject split, let $\mathcal{S}_{\mathrm{train}}$ denote the subjects in the training set. We minimize the empirical cross-entropy loss over the training subjects:
\begin{equation}
\mathcal{L}_{\mathrm{train}}(\theta) = \frac{1}{N_{\mathrm{train}}}
\sum_{s \in \mathcal{S}_{\mathrm{train}}} \sum_{i=1}^{N_s}
\ell(f_\theta(X_{s,i}),\, y_{s,i}), 
\label{eq:loss}
\end{equation}
where $N_{\mathrm{train}} = \sum_{s \in \mathcal{S}_{\mathrm{train}}} N_s$ and
$\ell(\cdot, \cdot)$ denotes the cross-entropy loss.

At iteration $t$, an optimizer samples a mini-batch $\mathcal{B}_t$ from the training split and computes the stochastic gradient
\begin{equation}
g_t = \nabla_\theta \mathcal{L}_{\mathcal{B}_t}(\theta_{t-1}),
\label{eq:stochastic_grad}
\end{equation}
where $\mathcal{L}_{\mathcal{B}_t}$ is the mini-batch loss. A generic adaptive optimizer then updates the parameters as
\begin{equation}
\theta_t = \theta_{t-1} - \eta_t \, \mathcal{U}_t(g_t, \mathcal{M}_{t-1}),
\label{eq:generic_update}
\end{equation}
where $\eta_t$ is the learning rate at step $t$, $\mathcal{U}_t(\cdot)$ is the optimizer-specific update direction, and $\mathcal{M}_{t-1}$ denotes the optimizer state accumulated from past steps.

To evaluate generalization on unseen subjects, we adopt a strict cross-subject protocol. The subjects are split into three subsets $\mathcal{S}_{\text{train}}, \mathcal{S}_{\text{val}}, \mathcal{S}_{\text{test}} \subseteq \{1,\dots,S\}$ satisfying $\mathcal{S}_{\text{train}} \cap \mathcal{S}_{\text{val}} = \mathcal{S}_{\text{val}} \cap \mathcal{S}_{\text{test}} = \mathcal{S}_{\text{train}} \cap \mathcal{S}_{\text{test}} = \emptyset$. The corresponding datasets are defined as
\begin{equation}
\mathcal{D}_{\mathrm{split}} = \bigcup_{s \in \mathcal{S}_{\mathrm{split}}} \mathcal{D}_s, \quad \mathrm{split} \in \{\mathrm{train}, \mathrm{val}, \mathrm{test}\}.
\label{eq:cross_subject_split}
\end{equation}
Because the subject sets are disjoint, the training and test splits are drawn from different subjects with different data distributions, i.e., \(\mathbb{P}_{\mathrm{train}} \neq \mathbb{P}_{\mathrm{test}}\). This setting therefore measures cross-subject generalization directly. Within the training split, however, mini-batches drawn from heterogeneous subjects can exhibit substantial variation in gradient statistics across iterations, making the optimization process nonstationary.

\subsection{Residual-Alignment Signal Scorer (RASS)}

AFOR sets the second-moment decay coefficient to a per-tensor, per-step value
$\beta_{2,t} \in [\beta_2^{\min}, \beta_2^{\text{init}}]$. Each parameter
tensor maintains its own scalar summary states throughout this process.
Training batches are drawn from a few subjects at a time, so $v_t$ averages
squared gradients over whichever subjects were sampled recently. With a large
$\beta_{2,t}$, the preconditioner stays weighted toward subjects that are no
longer sampled. Since $m_{t-1}$ is dominated by those subjects, a change of
sampled subject raises $e_t$ and lowers $\rho_t$.

For readability, we drop the tensor index $(k)$ in the per-tensor equations below and write $d_k$ for the number of elements in tensor $k$.

\textbf{Dual-timescale relative noise.} The instantaneous noise is the mean
absolute deviation between the current stochastic gradient $g_t$ and the
preceding first-moment estimate $m_{t-1}$, averaged over the $d_k$ elements
of the tensor:
\begin{equation}
e_t = \frac{1}{d_k} \sum_{j=1}^{d_k} \bigl| g_{t,j} - m_{t-1,j} \bigr|.
\label{eq:noise_inst}
\end{equation}
Two exponential moving averages track $e_t$ on distinct time scales:
\begin{equation}
\begin{aligned}
n_t^{\text{f}} &= \beta_1 \, n_{t-1}^{\text{f}} + (1-\beta_1) \, e_t, \\[2pt]
n_t^{\text{s}} &= \beta_2^{\text{init}} \, n_{t-1}^{\text{s}} + (1-\beta_2^{\text{init}}) \, e_t,
\end{aligned}
\label{eq:noise_dual}
\end{equation}
with $n_t^{\text{f}}$ fast (decay $\beta_1 = 0.9$) and $n_t^{\text{s}}$
slow (decay $\beta_2^{\text{init}} = 0.999$). We use the larger reference to obtain a conservative noise estimate. The slow channel provides a stable baseline, whereas the fast channel responds to noise spikes. As a result, the reference adapts to each tensor's own history without requiring an absolute threshold.

\textbf{Direction awareness.} The first moment follows the conventional
update $m_t = \beta_1 m_{t-1} + (1-\beta_1) g_t$. We define the
magnitude-to-residual ratio as its mean absolute value normalized by the noise
reference:
\begin{equation}
\rho_t^{m} = \frac{\frac{1}{d_k} \sum_{j=1}^{d_k} |m_{t,j}|}
                              {\max\!\bigl(n_t^{\text{f}},\; n_t^{\text{s}}\bigr) + \varepsilon}.
\label{eq:magnitude_ratio}
\end{equation}
A high $\rho_t^{m}$ means that the momentum magnitude is large relative to the
current noise reference. The temporal ordering is fixed throughout: $e_t$
and $\cos_t$ are evaluated against the pre-update momentum $m_{t-1}$. The
first moment is then advanced to $m_t$, and only this updated estimate
enters the magnitude term $\rho_t^{m}$.

A large momentum magnitude does not necessarily indicate reliable progress. The gradient direction may still oscillate across iterations. Therefore, we discount the score by a direction consistency factor. Let
\begin{equation}
\begin{aligned}
\bar{c}_t &= 0.9\, \bar{c}_{t-1} + 0.1 \cos_t, \\[2pt]
\cos_t &= \max\!\left(0,\; \frac{\langle g_t,\, m_{t-1} \rangle}
                          {\|g_t\|_2 \; \|m_{t-1}\|_2 + \varepsilon} \right),
\end{aligned}
\label{eq:dir}
\end{equation}
with $\bar{c}_0 = 1$. Here $\cos_t$ is the cosine similarity between the incoming gradient and the previous momentum direction, truncated at $0$. The temporal average $\bar{c}_t \in [0,1]$ approaches one under persistent alignment and $0$ under repeated disagreement.
The $\varepsilon$ term prevents division by $0$ when either vector is $0$. The resulting tensor-wise score is
\begin{equation}
\rho_t = \rho_t^{m} \cdot \bigl(1 + w \cdot (\bar{c}_t - 1)\bigr),
\qquad
\rho_t \leftarrow \max(\rho_t,\, 0),
\label{eq:score}
\end{equation}
where $w \geq 0$ is a direction weight (default $w = 1$): at $\bar{c}_t
= 1$ the score is unchanged, and below $1$ it shrinks in proportion to
the inconsistency. Both noisy gradients and unstable directions warrant
faster forgetting.

\subsection{Adaptive Forgetting Controller (AFC)}

The AFC maps $\rho_t$ onto the admissible range
$[\beta_2^{\min},\beta_2^{\text{init}}]$ and keeps the second-moment
estimator consistent as the decay coefficient changes at every step.

\textbf{Self-referential normalization.} The mapping from $\rho_t$ to $\beta_{2,t}$ must operate across heterogeneous tensor types whose score distributions differ in location and scale. Because score scales differ across tensors, each tensor uses its own online running statistics for normalization. Specifically, $\mu_t^{\rho}$ is the running mean of $\rho_t$, $\delta_t$ is the centered residual, and $\sigma_t^{\rho}$ is the running standard deviation:
\begin{equation}
\begin{aligned}
\mu_t^{\rho} &= \beta_2^{\text{init}} \mu_{t-1}^{\rho}
                + (1-\beta_2^{\text{init}}) \rho_t,
                \qquad \mu_0^{\rho} = 0, \\[2pt]
\delta_t &= \rho_t - \mu_t^{\rho}, \\[2pt]
(\sigma_t^{\rho})^{2} &= \beta_2^{\text{init}} (\sigma_{t-1}^{\rho})^{2}
                + (1-\beta_2^{\text{init}}) (\delta_t)^{2},
                \qquad (\sigma_0^{\rho})^{2} = 1.
\end{aligned}
\label{eq:zscore_stats}
\end{equation}
The standardized score is then computed from these centered and
scaled statistics:
\begin{equation}
\zeta_t = \frac{\delta_t}{\sqrt{(\sigma_t^{\rho})^{2} + \varepsilon}},
\qquad
\zeta_t \leftarrow \operatorname{clip}(\zeta_t,\; -5,\; 5).
\label{eq:zscore}
\end{equation}
A positive $\zeta_t$ indicates a score above its running mean and increases $\beta_{2,t}$. A negative value indicates a noisier update and decreases $\beta_{2,t}$. The clipping to $[-5, 5]$ prevents extreme values from saturating the sigmoid. Unlike schemes that compare the score to fixed thresholds, which require per-dataset or per-model tuning, the Z-score is scale-free and self-calibrating.

The raw adaptive coefficient is obtained by mapping $\zeta_t$ through a
logistic sigmoid into the admissible range:
\begin{equation}
\beta_{2,t}^{\text{raw}} = \beta_2^{\min}
        + (\beta_2^{\text{init}} - \beta_2^{\min}) \cdot \operatorname{sigmoid}(\zeta_t),
\label{eq:beta2_raw}
\end{equation}
where $\operatorname{sigmoid}(x) = 1 / (1 + e^{-x})$. The range $[0.99,0.999]$ corresponds to effective memory lengths of approximately 100-1000 steps. Larger coefficients retain longer histories, whereas smaller coefficients respond faster to gradient changes.

\textbf{Temporal stabilization gate (warm-up).} In early training, the noise estimates and Z-score statistics are unreliable because they are initialized to fixed constants (zero mean and unit variance). Allowing $\beta_{2,t}$ to deviate from $\beta_2^{\text{init}}$ during this phase can destabilize optimization. We therefore apply a linear warm-up gate.

\begin{equation}
\gamma_t = \min\!\left(1,\; \frac{t}{T_w}\right),\qquad
\beta_{2,t} = \gamma_t \cdot \beta_{2,t}^{\text{raw}}
            + (1 - \gamma_t) \cdot \beta_2^{\text{init}},
\label{eq:gate}
\end{equation}
where $T_w = 100$ steps. For $t<T_w$, the coefficient remains close to $\beta_2^{\text{init}}$, after the warm-up, the adaptive mechanism is fully applied.

\textbf{Generalized initialization correction.} The second-moment estimate uses the time-varying coefficient:
\begin{equation}
v_t = \beta_{2,t} \cdot v_{t-1} + (1 - \beta_{2,t}) \cdot g_t \odot g_t.
\label{eq:v_adaptive}
\end{equation}
Since $\beta_{2,t}$ changes across steps, the standard bias correction
$1 - \beta_2^{t}$ no longer applies.  The effective aggregate decay from
initialization to step $t$ is the cumulative product:
\begin{equation}
C_t = \prod_{s=1}^{t} \beta_{2,s} = C_{t-1} \cdot \beta_{2,t},
\qquad C_0 = 1.
\label{eq:cumprod}
\end{equation}
The initialization-corrected estimates and the parameter update (with decoupled weight decay
of strength $\lambda \geq 0$, following~\cite{adamw}) are:
\begin{equation}
\begin{aligned}
\hat{m}_t &= \frac{m_t}{1 - \beta_1^{t}}, \qquad
\hat{v}_t = \frac{v_t}{1 - C_t}, \\[6pt]
\theta_t &= \theta_{t-1}
           - \eta_t\lambda\theta_{t-1}
           - \eta_t \cdot \frac{\hat{m}_t}{\sqrt{\hat{v}_t} + \varepsilon}.
\end{aligned}
\label{eq:update}
\end{equation}
When $\beta_{2,t}$ is constant, $C_t$ reduces to $\beta_2^{t}$. With
decoupled weight decay, Eq.~\eqref{eq:update} then has the standard AdamW
form; setting $\lambda=0$ gives Adam. In the time-varying case, the
telescoping identity
$\sum_{i=1}^{t}(1-\beta_{2,i})\prod_{j=i+1}^{t}\beta_{2,j} = 1-C_t$ shows
that $1-C_t$ matches the total attenuation induced by the second-moment
recursion for a deterministic coefficient sequence. When the coefficient
responds to the gradient, this factor should be interpreted as an
initialization normalization rather than a guarantee of unbiasedness.

Algorithm~1 summarizes the AFOR update. RASS uses the uncorrected first
moment $m_t$ to compute the signal-quality score, whereas the parameter
update uses the bias-corrected moment $\hat m_t$. Local gradient statistics are
first converted into a tensor-wise signal-quality score, then mapped to a
bounded second-moment decay coefficient, and finally applied through
time-varying second-moment updates with cumulative-product normalization.

\begin{algorithm}[!ht]
\caption{AFOR update skeleton. All tensor-wise states are maintained per
parameter tensor. $\operatorname{mean}(\cdot)$ averages tensor elements;
$\operatorname{sigmoid}(\cdot)$ is the logistic sigmoid.}
\label{alg:afor}
\footnotesize
\begin{algorithmic}[1]
\REQUIRE learning rate $\eta_t$; momentum decay $\beta_1$; slow decay
$\beta_2^{\text{init}}$; forgetting lower bound $\beta_2^{\min}$;
direction weight $w$; warm-up steps $T_w$; weight decay $\lambda$;
stability constant $\varepsilon$
\STATE \textbf{Initialize} per-tensor states for each parameter tensor $k$:
$m_0^{(k)}=v_0^{(k)}=\mathbf{0}$, $n_0^{\text{f},(k)}=n_0^{\text{s},(k)}=0$,
  $\bar{c}_0^{(k)}=1$, $\mu_0^{\rho,(k)}=0$, $(\sigma_0^{\rho,(k)})^{2}=1$,
$C_0^{(k)}=1$
\FOR{$t = 1, 2, \dots$}
  \FOR{each parameter tensor $k$}
  \STATE $g_t^{(k)} \leftarrow \nabla_{\theta^{(k)}}\mathcal{L}_{\mathcal{B}_t}(\theta_{t-1})$
  \STATE \textbf{RASS:} score local gradient quality from residual noise and direction consistency.
  \STATE $e_t^{(k)} \leftarrow \operatorname{mean}\bigl(|g_t^{(k)} - m_{t-1}^{(k)}|\bigr)$;\;
         update $n_t^{\text{f},(k)}$, $n_t^{\text{s},(k)}$, and $\bar{c}_t^{(k)}$
  \STATE $m_t^{(k)} \leftarrow \beta_1 m_{t-1}^{(k)} + (1-\beta_1) g_t^{(k)}$
  \STATE $\rho_t^{m,(k)} \leftarrow \operatorname{mean}\bigl(|m_t^{(k)}|\bigr) \big/ \bigl(\max(n_t^{\text{f},(k)},\, n_t^{\text{s},(k)}) + \varepsilon\bigr)$
  \STATE \textbf{AFC:} map the score to a bounded second-moment decay coefficient and update the initialization-correction factor.
  \STATE $\rho_t^{(k)} \leftarrow \max\!\bigl(0,\rho_t^{m,(k)}(1+w(\bar{c}_t^{(k)}-1))\bigr)$
  \STATE $\beta_{2,t}^{(k)} \leftarrow$ map $\rho_t^{(k)}$ to a bounded second-moment decay coefficient with warm-up gating
  \STATE $C_t^{(k)} \leftarrow C_{t-1}^{(k)}\cdot\beta_{2,t}^{(k)}$
  \STATE $v_t^{(k)} \leftarrow \beta_{2,t}^{(k)}\, v_{t-1}^{(k)} + (1-\beta_{2,t}^{(k)})\, g_t^{(k)} \odot g_t^{(k)}$
  \STATE \textbf{Update:} apply initialization correction and decoupled weight decay.
  \STATE $\hat{m}_t^{(k)} \leftarrow m_t^{(k)} \big/ (1-\beta_1^{t})$;\quad
  $\hat{v}_t^{(k)} \leftarrow v_t^{(k)} \big/ (1-C_t^{(k)})$
  \STATE $\theta_t^{(k)} \leftarrow (1-\eta_t\lambda)\,\theta_{t-1}^{(k)}
    - \eta_t\, \hat{m}_t^{(k)} \big/ \bigl(\sqrt{\hat{v}_t^{(k)}} + \varepsilon\bigr)$
  \ENDFOR
\ENDFOR
\end{algorithmic}
\end{algorithm}

\section{IMPLEMENTATION DETAILS}

\subsection{Dataset and Preprocessing}

We evaluate on three EEG decoding benchmarks covering emotion recognition (ER), motor imagery (MI), and sleep staging (SS). Table~\ref{tab:datasets} summarizes their basic information.

\textbf{DEAP}~\cite{2011deap} records EEG at 128\,Hz from 32 subjects, with each one watching 40 one-minute videos and providing continuous valence and arousal ratings
on a 1-9 scale. We binarize valence at threshold~5, following standard practice~\cite{libeer}. A 3\,s
pre-trial baseline is subtracted from each trial, and a fifth-order Butterworth
bandpass filter (0.3-50\,Hz) is applied. Each 60\,s trial is segmented into
non-overlapping 1s clips of 128 time points.

\textbf{BCI Competition IV 2a}~\cite{bci42a} is a MI dataset covering
left hand, right hand, feet, and tongue tasks, recorded at 250\,Hz from 9 subjects over 22 EEG
channels. Following
the benchmark protocol, we extract the interval 0.5-3.5\,s after cue onset
(750 samples) from the raw recordings. The same fifth-order Butterworth
bandpass filter (0.3-50\,Hz) as for DEAP is applied, and each 750-point trial
is then segmented into overlapping 250-point windows with a 125-point stride.

\textbf{ISRUC-Sleep SG-I}~\cite{isruc} contains polysomnography recorded at
200\,Hz from 100 subjects over six EEG channels, covering sleep stages Wake, N1, N2, N3, and REM. No bandpass filtering is applied to this dataset. Each recording is
partitioned into 30\,s epochs of 6000 time points, which are segmented into
overlapping 600-point windows with a 300-point stride.
\subsection{Splitting and Cross-validation}

All experiments follow a strict cross-subject protocol with disjoint training,
validation, and test subjects. DEAP and BCI IV use leave-one-subject-out
cross-validation (LOSO-CV), where each round holds out one subject as the test
set $\mathcal{D}_{\mathrm{test}}$ and every subject is tested once. Because ISRUC contains 100 subjects, each round uses ten test subjects, so that every
subject is still tested once across the ten rounds. From the remaining subjects, approximately 20\% are randomly selected as the validation set $\mathcal{D}_{\mathrm{val}}$, with the rest forming the training set $\mathcal{D}_{\mathrm{train}}$. For each fold, the validation split is drawn five times with different random seeds to reduce the effect of the random validation-subject selection; each fold's result is the average of five times, and performance is reported as the fold-wise mean $\pm$ standard deviation ($std$) across folds of test-set accuracy (\textit{Acc}) and weighted F1 (\textit{wF1}).
Table~\ref{tab:datasets} lists the details of the data division.

\begin{table}[t]
\centering
\caption{Basic information and per-round subject counts of the three datasets under the cross-subject protocol.}
\label{tab:datasets}
\small
\renewcommand{\arraystretch}{1.08}
\setlength{\tabcolsep}{5pt}
\begin{tabular*}{\columnwidth}{@{\extracolsep{\fill}}lcccccc}
\hline
\textbf{Dataset} & \textbf{Task} & \textbf{Channels} & \textbf{Classes} & \textbf{Train} & \textbf{Val.} & \textbf{Test} \\
\hline
DEAP & ER & 32 & 2 & 25 & 6 & 1 \\
BCI IV & MI & 22 & 4 & 6 & 2 & 1 \\
ISRUC & SS & 6 & 5 & 70 & 20 & 10 \\
\hline
\end{tabular*}
\end{table}

\begin{table}[t]
\centering
\caption{Selected learning rate for each backbone-dataset pair.}
\label{tab:lr}
\small
\renewcommand{\arraystretch}{1.08}
\setlength{\tabcolsep}{5pt}
\begin{tabular*}{\columnwidth}{@{\extracolsep{\fill}}lccc}
\hline
\textbf{Backbone} & \textbf{DEAP} & \textbf{BCI IV 2a} & \textbf{ISRUC} \\
\hline
EEGNet & $1\times10^{-3}$ & $3\times10^{-3}$ & $3\times10^{-4}$ \\
ACRNN & $1\times10^{-3}$ & $3\times10^{-4}$ & $1\times10^{-4}$ \\
NSAL-DGAT & $1\times10^{-3}$ & $3\times10^{-4}$ & $3\times10^{-4}$ \\
HSLT & $1\times10^{-3}$ & $3\times10^{-4}$ & $3\times10^{-4}$ \\
\hline
\end{tabular*}
\end{table}

\begin{figure}[t]
\centering
\includegraphics[width=\columnwidth]{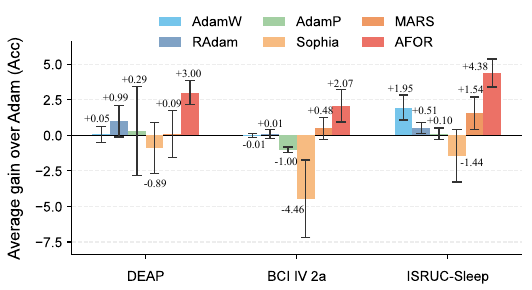}
\caption{Average \textit{Acc} gain over Adam on each dataset. Each bar shows the average gain of different optimizers compared to Adam across EEGNet, ACRNN, NSAL-DGAT, and HSLT, and the error bars indicate the standard deviation across the four backbones.}
\label{fig:acc_gain}
\end{figure}

\subsection{Experiment Settings}

All experiments were run on four NVIDIA RTX 3090 Ti GPUs using
Python 3.10.18 and the PyTorch 2.7.1 framework with CUDA 12.6 support.
AFOR is implemented in PyTorch and released as an open-source package. The
optimizer can be installed with \texttt{pip install afor-optimizer}.

\noindent\textbf{Learning rate selection.}
A single learning rate is not suitable for all optimizer-model-dataset
combinations. To isolate the effect of the update rule, we first evaluate all optimizers under a shared learning-rate protocol. The learning rate is selected using Adam for each backbone-dataset pair and then fixed across optimizers. For each combination, we run a grid search over
$\{1\times10^{-4}, 3\times10^{-4}, 1\times10^{-3}, 3\times10^{-3},
1\times10^{-2}\}$ using the Adam optimizer. To maintain efficient searching, the grid runs use one cross-validation
round, 50 training epochs, and early stopping with a patience
of 15 epochs. The learning rate yielding the highest validation \textit{Acc} is selected
and fixed on all experiments. The search results are shown in Table~\ref{tab:lr}.

\noindent\textbf{Training configuration.}
All models are trained with cross-entropy loss and early stopping, with
patience 30 epochs and a maximum of 200 epochs. The batch size is 256 for
DEAP and 64 for BCI IV and ISRUC. All Adam-family baselines share
$\beta_1 = 0.9$, $\beta_2 = 0.999$, and $\varepsilon = 10^{-8}$, while
Sophia uses its published defaults ($\beta_1 = 0.965$, $\beta_2 = 0.99$,
 $\rho_{\mathrm{Sophia}} = 0.04$, $K_{\mathrm{Sophia}} = 10$). To keep the comparison fair, all optimizers use
the same $\texttt{weight\_decay}=1\times10^{-4}$. The remaining algorithm-specific
hyperparameters keep their published default values. No data augmentation is
applied beyond the preprocessing and segmentation described above. AFOR
inherits the same $\beta_1$,
$\beta_2^{\text{init}}$, and $\varepsilon$, with the other parameters fixed at
the design values reported in Section~III.
For each selected model-dataset pair, the experimental learning-rate schedule
is constant within a run, i.e., $\eta_t\equiv\eta$.

\begin{table*}[t]
\centering
\caption{Results on three EEG benchmarks under the cross-subject protocol.}
\label{tab:results}
\footnotesize
\renewcommand{\arraystretch}{1.08}
\setlength{\tabcolsep}{5pt}
\begin{tabular*}{\textwidth}{@{\extracolsep{\fill}}lcccccccc}
\toprule
\multirow{2}{*}{\textbf{Method}}& \multicolumn{2}{c}{\textbf{EEGNet}} & \multicolumn{2}{c}{\textbf{ACRNN}}
& \multicolumn{2}{c}{\textbf{NSAL-DGAT}} & \multicolumn{2}{c}{\textbf{HSLT}} \\
\cmidrule(lr){2-3}\cmidrule(lr){4-5}\cmidrule(lr){6-7}\cmidrule(lr){8-9}
 & \textit{Acc} & \textit{wF1} & \textit{Acc} & \textit{wF1}
& \textit{Acc} & \textit{wF1} & \textit{Acc} & \textit{wF1} \\
\midrule
\multicolumn{9}{c}{\textbf{DEAP \quad \emph{Emotion recognition (2-class)}}} \\
\midrule
Adam    & 60.90$\pm$7.48 & 52.58$\pm$9.79 & 59.32$\pm$8.76 & 47.13$\pm$12.46 & 61.59$\pm$6.57 & 52.34$\pm$10.26 & 56.99$\pm$7.65 & 47.62$\pm$8.02 \\
AdamW   & 60.08$\pm$7.26 & 51.86$\pm$10.67 & 59.81$\pm$8.27 & 47.92$\pm$12.11 & 62.17$\pm$7.65 & 52.63$\pm$9.68 & 56.96$\pm$7.67 & 47.72$\pm$8.13 \\
RAdam   & \underline{62.66$\pm$8.47$^{*}$} & \underline{54.08$\pm$9.72$^{*}$} & \underline{61.72$\pm$8.13$^{*}$} & \underline{51.61$\pm$12.81$^{*}$} & 61.37$\pm$8.86 & 52.20$\pm$10.31 & 57.02$\pm$7.81 & 46.02$\pm$8.34 \\
AdamP   & 60.43$\pm$8.06 & 50.88$\pm$10.03 & 55.31$\pm$9.14 & 42.51$\pm$10.78 & \underline{62.48$\pm$8.32$^{*}$} & \underline{54.29$\pm$9.71$^{*}$} & \cellcolor{gray!40}61.73$\pm$6.29$^{*}$ & \cellcolor{gray!40}52.32$\pm$7.56$^{*}$ \\
Sophia  & 61.72$\pm$8.42$^{*}$ & 52.74$\pm$11.63 & 55.94$\pm$8.74 & 40.53$\pm$10.40 & 59.73$\pm$10.04 & 48.49$\pm$12.58 & 57.84$\pm$7.26 & 50.58$\pm$9.07$^{*}$ \\
MARS    & 60.03$\pm$8.64 & 52.35$\pm$9.81 & 58.71$\pm$8.28 & 46.98$\pm$11.99 & 60.46$\pm$7.51 & 51.78$\pm$10.13 & 59.95$\pm$8.12$^{*}$ & 49.16$\pm$9.35$^{*}$ \\
\textbf{AFOR(Ours)} & \cellcolor{gray!40}62.94$\pm$7.25$^{*}$ & \cellcolor{gray!40}54.59$\pm$9.34$^{*}$ & \cellcolor{gray!40}62.86$\pm$7.80$^{*}$ & \cellcolor{gray!40}54.52$\pm$8.72$^{*}$ & \cellcolor{gray!40}63.92$\pm$7.67$^{*}$ & \cellcolor{gray!40}54.45$\pm$9.74$^{*}$ & \underline{61.08$\pm$7.14$^{*}$} & \underline{52.17$\pm$7.70$^{*}$} \\
\midrule
\multicolumn{9}{c}{\textbf{BCI Competition IV 2a \quad \emph{Motor imagery (4-class)}}} \\
\midrule
Adam    & \underline{37.22$\pm$2.45} & 33.63$\pm$3.27 & 36.55$\pm$3.83 & 29.94$\pm$3.92 & 37.80$\pm$3.18 & 34.96$\pm$3.47 & 39.54$\pm$2.75 & 35.88$\pm$2.95 \\
AdamW   & 37.14$\pm$2.72 & 34.94$\pm$3.73$^{*}$ & 36.35$\pm$4.13 & 27.51$\pm$4.44 &37.97$\pm$3.30 & 34.34$\pm$3.40 & 39.59$\pm$2.01 & 35.52$\pm$3.56 \\
RAdam   & 37.22$\pm$2.74 & \underline{35.25$\pm$3.29$^{*}$} & 36.79$\pm$4.23 & 30.23$\pm$3.41 & 38.30$\pm$2.39 & 34.45$\pm$3.80 & 39.15$\pm$2.38 & 35.89$\pm$4.02 \\
AdamP   & 36.12$\pm$2.59 & 33.45$\pm$4.11 & 35.47$\pm$3.76 & \underline{30.42$\pm$3.65} & 37.12$\pm$4.40 & 32.69$\pm$4.16 & 38.38$\pm$4.04 & 33.29$\pm$2.37 \\
Sophia  & 28.06$\pm$1.01 & 16.74$\pm$5.39 & 34.07$\pm$4.71 & 24.23$\pm$4.00 & 34.46$\pm$5.52 & 30.65$\pm$6.23 &36.68$\pm$3.22 & 30.03$\pm$5.97 \\
MARS    & 36.57$\pm$2.52 & 34.40$\pm$3.61$^{*}$ & \underline{36.83$\pm$3.78} & 30.22$\pm$4.30 & \underline{38.58$\pm$3.38$^{*}$} & \underline{35.47$\pm$4.03} & \cellcolor{gray!40}41.05$\pm$2.58$^{*}$ & \cellcolor{gray!40}37.75$\pm$2.24$^{*}$ \\
\textbf{AFOR(Ours)} & \cellcolor{gray!40}37.94$\pm$2.18 & \cellcolor{gray!40}36.21$\pm$3.47$^{*}$ & \cellcolor{gray!40}40.26$\pm$3.24$^{*}$ & \cellcolor{gray!40}35.27$\pm$3.52$^{*}$ & \cellcolor{gray!40}40.29$\pm$3.72$^{*}$ & \cellcolor{gray!40}36.79$\pm$3.03$^{*}$ & \underline{40.91$\pm$1.89} & \underline{37.53$\pm$3.04$^{*}$} \\
\midrule
\multicolumn{9}{c}{\textbf{ISRUC-Sleep SG-I \quad \emph{Sleep staging (5-stage)}}} \\
\midrule
Adam    & 62.06$\pm$3.01 & 58.69$\pm$3.15 & 64.63$\pm$3.61 & 60.93$\pm$5.26 & 64.58$\pm$3.04 & 60.71$\pm$4.47 & 63.50$\pm$3.67 & 58.38$\pm$5.71 \\
AdamW   & \underline{64.48$\pm$3.72$^{*}$} & \underline{60.42$\pm$4.17$^{*}$} & \underline{67.59$\pm$3.67$^{*}$} & 62.00$\pm$4.72$^{*}$ & \underline{66.41$\pm$2.62$^{*}$} & 61.88$\pm$5.50 & 64.08$\pm$3.33 & 58.96$\pm$4.14 \\
RAdam   & 62.39$\pm$4.52 & 59.10$\pm$4.76 & 64.61$\pm$3.92 & 60.89$\pm$4.36 & 65.62$\pm$3.42$^{*}$ & 60.62$\pm$4.20 & 64.17$\pm$4.25 & 58.67$\pm$5.92 \\
AdamP   & 62.41$\pm$4.25 & 56.53$\pm$4.97 & 64.25$\pm$4.73 & 60.51$\pm$5.36 & 65.23$\pm$3.28 & 61.59$\pm$5.33 & 63.29$\pm$3.92 & 57.67$\pm$6.08 \\
Sophia  & 61.09$\pm$5.77 & 54.20$\pm$5.32 & 60.60$\pm$5.36 & 55.12$\pm$6.76 & 65.71$\pm$3.58$^{*}$ & 63.13$\pm$4.27$^{*}$ & 61.62$\pm$5.12 & 54.79$\pm$8.50 \\
MARS    & 62.55$\pm$3.91 & 57.50$\pm$4.40 & 65.28$\pm$4.52 & \underline{62.03$\pm$5.47$^{*}$} & 66.24$\pm$2.55$^{*}$ & \underline{63.26$\pm$3.34$^{*}$}& \cellcolor{gray!40}66.85$\pm$3.21$^{*}$ & \underline{61.45$\pm$4.38}$^{*}$ \\
\textbf{AFOR(Ours)} & \cellcolor{gray!40}67.57$\pm$3.12$^{*}$ & \cellcolor{gray!40}64.33$\pm$4.54$^{*}$ & \cellcolor{gray!40}68.92$\pm$3.31$^{*}$ & \cellcolor{gray!40}64.97$\pm$5.01$^{*}$ & \cellcolor{gray!40}69.46$\pm$2.96$^{*}$ & \cellcolor{gray!40}64.83$\pm$4.16$^{*}$ & \underline{66.35$\pm$2.71$^{*}$} & \cellcolor{gray!40}62.42$\pm$4.33$^{*}$ \\
\bottomrule
\end{tabular*}
\vspace{2pt}
\parbox{\textwidth}{\footnotesize $^{*}$ Significantly higher than Adam under a one-sided paired permutation test with Holm correction ($p<0.05$).}
\end{table*}

\subsection{Backbones and Baselines}

\noindent\textbf{Backbones.}
To evaluate the optimizer on different architectures, we select four representative backbones that cover the main paradigms in EEG applications: \textbf{EEGNet}~\cite{eegnet} (CNN-based) with depthwise and separable convolutions, \textbf{ACRNN}~\cite{acrnn} (RNN-based) with convolutional layers and recurrent units, \textbf{NSAL-DGAT}~\cite{nsaldgat} (GNN-based) with dynamic graph attention over EEG channels, and \textbf{HSLT}~\cite{hslt} (Transformer-based) combines convolutional tokenization with hierarchical self-attention for EEG decoding.

\begin{figure*}[!t]
\centering
\includegraphics[width=\textwidth]{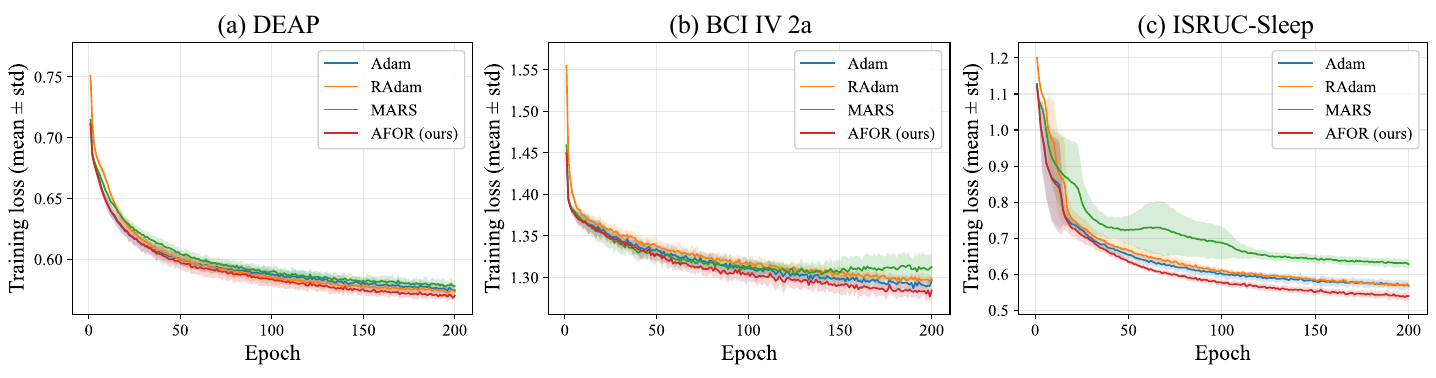}
\caption{Training-loss trajectories of AFOR against Adam, RAdam, and MARS on the EEGNet backbone over 200 epochs without early stopping. Figs (a), (b), and (c) correspond to DEAP, BCI IV, and ISRUC, respectively.}
\label{fig:convergence}
\end{figure*}

\noindent\textbf{Baselines.}
We compare AFOR against six representative optimizers spanning four
families of adaptive methods, including the de facto defaults in EEG
decoding.  \textbf{Adam} and \textbf{AdamW} represent standard
first-order adaptive methods based on moment preconditioning, the latter
decoupling weight decay from the update.  \textbf{RAdam} and
\textbf{AdamP} represent corrections of the adaptive step size,
rectifying the early-phase learning-rate variance and stabilizing
scale-invariant weights, respectively.  \textbf{Sophia} represents
second-order methods through a lightweight diagonal Hessian
preconditioner, and \textbf{MARS} represents gradient variance reduction
for noisy settings.  Across these families, however, the decay coefficient of
the second-moment estimate remains a fixed global constant, imposing one
forgetting schedule on all parameters and all training stages.  AFOR
adapts this coefficient per tensor and per step, a dimension no
baseline covers.

\section{EXPERIMENT}

\subsection{Main Results}

Table~\ref{tab:results} reports the fold-wise mean$\pm std$ deviation of classification \textit{Acc} and \textit{wF1}, where each fold's value averages its five validation-set draws and the statistics are computed across folds. 
The best result within each dataset-backbone block is shaded in gray and the second-best is
\underline{underlined}. Averaged over the twelve backbone-dataset
combinations, AFOR improves upon Adam, the default optimizer of the field, by
3.15\% in \textit{Acc} (up to 5.51\%) and by 3.77\% on \textit{wF1} (up to 7.39\%), with consistent gains across all backbones and datasets.
We also performed one-sided paired permutation tests on the fold-level results
to identify improvements over Adam. After Holm correction~\cite{Holm1979ASS} across the six optimizer comparisons, AFOR shows significant improvements over Adam in 22/24 backbone-metric combinations across the three datasets, better than the other compared optimizer. In Table~\ref{tab:results}, $^*$ denotes a significant improvement over Adam after a one-sided paired permutation test with Holm correction ($p<0.05$).

Fig.~\ref{fig:acc_gain} further summarizes Table~\ref{tab:results} from a
dataset-level view. AFOR gives the largest positive mean gain on all three
benchmarks. The largest dataset-level gain
appears on ISRUC, which may reflect that
adaptive forgetting is more helpful when cross-subject nonstationarity and
label complexity are stronger. We note that AFOR does not reach the best
result on HSLT. A possible reason is that layer-normalized Transformer blocks reshape gradient magnitudes across layers, while tensor-wise forgetting estimates computed from very small per-region tensors may have a lower signal-to-noise ratio, making the resulting forgetting signal less distinctive \cite{li2025mixln,gray2024normalization}. Even so, AFOR still improves over Adam by 1.37-4.09\% \textit{Acc} across the three settings.

\subsection{Visualization}

\begin{figure*}[!t]
\centering
\includegraphics[width=\textwidth]{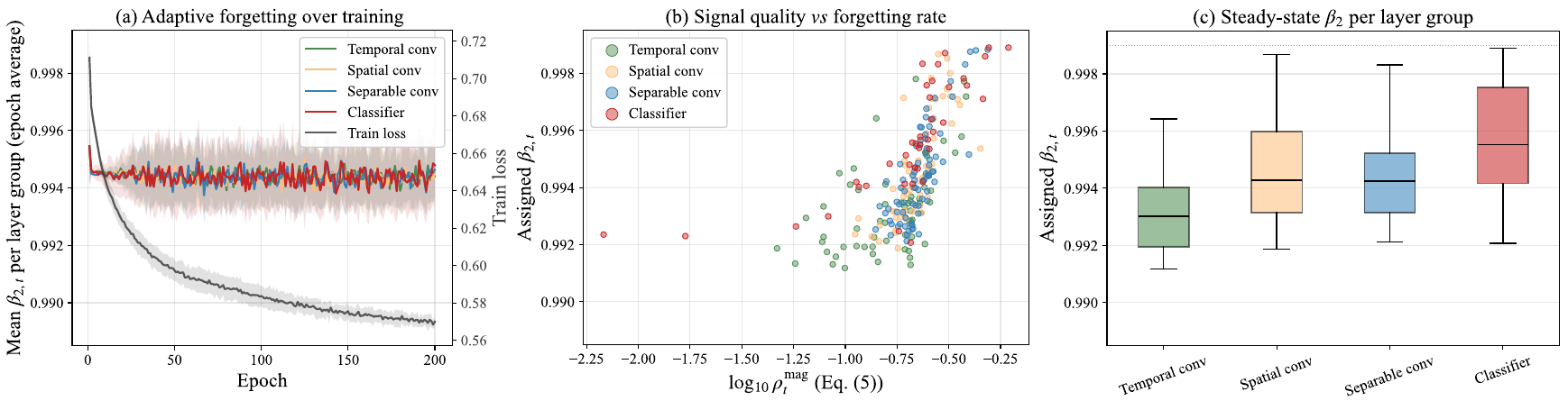}
\caption{Summary of the adaptive forgetting analysis on DEAP with the
EEGNet backbone. (a) epoch-averaged $\beta_{2,t}$ per layer group with
the within-epoch standard deviation and training loss; (b) magnitude-to-residual
ratio \textit{vs}. $\beta_{2,t}$, colored by layer group; (c)
steady-state $\beta_2$ distribution per layer group.}
\label{fig:analysis_summary}
\end{figure*}

\begin{table*}[t]
\centering
\caption{Modular ablations on DEAP and BCI IV 2a under LOSO-CV.}
\label{tab:ablation}
\footnotesize
\renewcommand{\arraystretch}{1.08}
\setlength{\tabcolsep}{5pt}
\begin{tabular*}{\textwidth}{@{\extracolsep{\fill}}llcccccccc}
\toprule
\multirow{2}{*}{\textbf{Module}} & \multirow{2}{*}{\textbf{Variant}} & \multicolumn{2}{c}{\textbf{DEAP-EEGNet}} & \multicolumn{2}{c}{\textbf{DEAP-ACRNN}}
& \multicolumn{2}{c}{\textbf{BCI-EEGNet}} & \multicolumn{2}{c}{\textbf{BCI-ACRNN}} \\
\cmidrule(lr){3-4}\cmidrule(lr){5-6}\cmidrule(lr){7-8}\cmidrule(lr){9-10}
 &  & \textit{Acc} & \textit{wF1} & \textit{Acc} & \textit{wF1}
& \textit{Acc} & \textit{wF1} & \textit{Acc} & \textit{wF1} \\
\midrule
\textbf{AFOR} & -- & 62.94$\pm$7.25 & 54.59$\pm$9.34 & 62.86$\pm$7.80 & 54.52$\pm$8.72 & 37.94$\pm$2.18 & 36.21$\pm$3.47 & 40.26$\pm$3.24 & 35.27$\pm$3.52 \\
\addlinespace
\multirow{2}{*}{\textbf{RASS}} 
& \textit{w/o} DC & 61.95$\pm$8.49 & 53.37$\pm$10.04 & 60.39$\pm$8.02 & 51.45$\pm$9.43 & 36.34$\pm$2.09 & 33.46$\pm$3.49 & 38.78$\pm$4.31 & 33.43$\pm$4.15 \\
 & \textit{w/o} FC & 62.14$\pm$7.32 & 53.85$\pm$8.42 & 62.55$\pm$8.56 & 54.12$\pm$9.29 & 37.64$\pm$2.00 & 35.39$\pm$3.53 & 39.84$\pm$3.49 & 34.67$\pm$3.90 \\
\addlinespace
\multirow{3}{*}{\textbf{AFC}} 
 & \textit{w/o} SRN & 61.93$\pm$8.26 & 53.44$\pm$9.80 & 61.08$\pm$9.27 & 51.89$\pm$10.45 & 37.03$\pm$2.70 & 34.14$\pm$3.98 & 37.26$\pm$4.48 & 33.26$\pm$3.98 \\
 & \textit{w/o} TSG & 62.55$\pm$9.57 & 54.23$\pm$10.05 & 62.33$\pm$7.83 & 53.82$\pm$8.62 & 37.82$\pm$2.59 & 35.93$\pm$3.76 & 39.84$\pm$3.26 & 34.87$\pm$3.75 \\
 & \textit{w/o} EC & 61.57$\pm$8.24 & 52.63$\pm$9.18 & 61.01$\pm$9.41 & 52.16$\pm$9.85 & 36.87$\pm$2.91 & 34.55$\pm$3.31 & 39.17$\pm$3.94 & 34.64$\pm$3.60 \\
\addlinespace
\multirow{3}{*}{\textbf{Fixed decay}}
 & $\beta_2{=}0.99$ & 61.47$\pm$7.28 & 52.61$\pm$9.23 & 61.45$\pm$8.25 & 52.54$\pm$10.46 & 36.94$\pm$2.70 & 34.82$\pm$3.64 & 38.58$\pm$3.41 & 33.40$\pm$4.54 \\
 & $\beta_2{=}0.999$ & 62.33$\pm$7.05 & 53.81$\pm$9.38 & 61.95$\pm$8.10 & 53.15$\pm$9.19 & 37.29$\pm$2.62 & 35.43$\pm$3.70 & 39.41$\pm$4.11 & 34.69$\pm$3.42 \\
\bottomrule
\end{tabular*}
\end{table*}

\begin{table}[t]
\centering
\caption{Trajectory statistics on the projected loss landscape of ISRUC-Sleep with EEGNet.}
\label{tab:landscape_metrics}
\footnotesize
\setlength{\tabcolsep}{8pt}
\begin{tabular}{lcc}
\toprule
\textbf{Method} & \textbf{Minimum projected loss} & \textbf{Projected path length} \\
\midrule
Adam & 3.427 & 48.947 \\
AFOR & 0.548 & 40.887 \\
\bottomrule
\end{tabular}
\end{table}

Next, we examine why these gains arise by following the optimization process
through some views, including the loss landscape, the layer-wise forgetting
behavior, and the convergence trajectory. The sensitivity of the main
hyperparameters is examined separately in Section~V-D.

The right of Fig.~\ref{fig:method} shows the loss landscape of ISRUC-EEGNet, spanned by two random, Gram-Schmidt-orthogonalized directions. Table~\ref{tab:landscape_metrics} quantifies the two trajectories, showing that AFOR reaches a lower projected loss while following a shorter path than Adam.
These measurements indicate more direct optimization progress in the projected landscape.

Fig.~\ref{fig:analysis_summary} traces the forgetting policy that AFOR
learns on DEAP-EEGNet. In the steady state, early convolutional groups
settle near the lower decay bound while the classifier retains a long
memory, and larger magnitude-to-residual scores are consistently mapped to
larger coefficients. This allocation matches each layer's role under
subject shift, since feature extractors face fast-changing gradients and
profit from short memory while the classifier accumulates stable
statistics and profits from long memory. The layer-wise pattern is produced online by the controller rather than specified in advance. The same design values were used for all four backbones.

Fig.~\ref{fig:convergence} shows how this policy affects the optimization trajectory. On all three datasets, AFOR tracks Adam, RAdam, and MARS through the early phase and keeps descending after they plateau. 
AFOR does not show a clear early-stage speed advantage. Its training loss continues to decrease after the fixed-decay baselines begin to plateau.

\subsection{Ablation Study}
Table~\ref{tab:ablation} reports modular ablations on DEAP and BCI IV
under LOSO-CV. The first column groups the rows by component family, and
the second column lists the removed operation or the control condition. DC stands for direction discount, FC for fast channel, SRN for self-referential normalization, TSG for temporal stabilization gate, and EC for cumulative-product correction. The last block replaces the adaptive
schedule with two fixed-decay controls, $\beta_2=0.99$ and $\beta_2=0.999$,
to test whether the gains come from adaptivity rather than a better constant
coefficient. When an operation is removed, the quantity it produced is
replaced as follows: \textit{w/o DC} sets the direction weight to $w=0$,
so the score reduces to the magnitude term $\rho_t^{m}$ alone;
\textit{w/o FC} drops the fast noise channel and uses the slow estimate
$n_t^{\text{s}}$ alone as the noise reference; \textit{w/o SRN} replaces
the self-referential Z-score with a fixed sigmoid mapping
$\operatorname{sigmoid}\bigl((\rho_t-\mu_{\mathrm{fix}})/s_{\mathrm{fix}}\bigr)$ with frozen constants $\mu_{\mathrm{fix}}=1.0$ and
$s_{\mathrm{fix}}=2.0$; \textit{w/o TSG} sets the gate $\gamma_t \equiv 1$, engaging the full adaptive decay from the first step; and \textit{w/o EC} replaces the cumulative-product correction $1-C_t$ with the standard fixed-decay
correction $1-(\beta_2^{\text{init}})^{t}$. All other settings match the full AFOR.

Within \textbf{RASS}, DC is the most important term. Removing it drops
\textit{Acc} by 0.99\% and 2.47\% on DEAP-EEGNet and DEAP-ACRNN, and
by 1.60\% and 1.48\% on BCI IV-EEGNet and BCI IV-ACRNN, respectively; the
corresponding \textit{wF1} drops are 1.22\%, 3.07\%, 2.75\%, and 1.84\%.
FC is milder but still consistent, with smaller drops of 0.30-0.80\%
\textit{Acc} and 0.40-0.82\% \textit{wF1} across the four
settings. The larger drop after removing DC indicates that directional consistency has a stronger effect than the fast noise channel in these settings.

Within \textbf{AFC}, SRN contributes the largest gain, especially on
the harder ACRNN settings. Removing SRN reduces \textit{Acc} by 1.01\% and
1.78\% on DEAP-EEGNet and DEAP-ACRNN, and by 0.91\% and 3.00\% on
BCI IV-EEGNet and BCI IV-ACRNN; the \textit{wF1} drops are 1.15\%, 2.63\%, 2.07\%,
and 2.01\%. TSG changes performance only slightly, with \textit{Acc}
decreases of 0.12-0.53\%, indicating that warm-up mainly stabilizes
the earliest updates. EC remains consistently useful, with \textit{Acc}
drops of 1.07-1.85\% and \textit{wF1} drops of 0.63-2.36\%,
because the cumulative-product correction keeps the time-varying
second-moment estimate properly normalized.

The two fixed-decay rows test whether a constant coefficient can reproduce the performance of the full adaptive schedule. $\beta_2=0.999$ is consistently
better than $\beta_2=0.99$, but both remain below full AFOR across all settings. The gaps range from 0.61\% to 1.68\% in \textit{Acc} and
from 0.58\% to 1.98\% in \textit{wF1}, so the gain is not explained by a
single fixed decay choice.

Overall, the ablations show that all components contribute, with DC, SRN, and EC providing the largest gains and FC and TSG playing smaller supporting roles. On DEAP and BCI IV-ACRNN, every ablated variant remains above its corresponding Adam baseline. The only exceptions occur on BCI IV-EEGNet, where removing DC, SRN, or EC falls slightly below Adam (36.34\%, 37.03\%,
and 36.87\% \textit{vs.} 37.22\%), and this is also the setting where full AFOR has its smallest margin over Adam ($+0.72\%$). Across all reported settings, full AFOR improves over the ablated variants by up to 3.00\% \textit{Acc} and 3.07\% \textit{wF1}.

\begin{figure*}[!t]
\centering
\includegraphics[width=\textwidth]{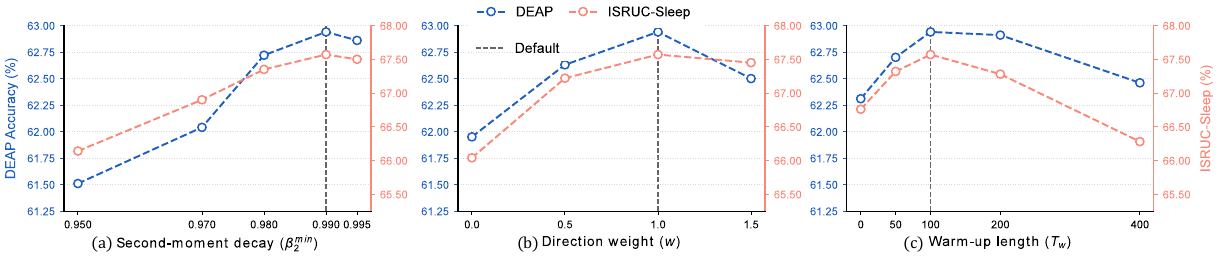}
\caption{Sensitivity analysis of AFOR on DEAP and ISRUC with the EEGNet
backbone. The dashed vertical line marks the default setting; the left and
right y-axes correspond to DEAP and ISRUC, respectively.}
\label{fig:sensitivity}
\end{figure*}

\begin{table*}[t]
\centering
\caption{Single-step optimizer cost comparison on DEAP with EEGNet.}
\label{tab:complexity}
\footnotesize
\renewcommand{\arraystretch}{1.08}
\setlength{\tabcolsep}{5pt}
\begin{tabular*}{\textwidth}{@{\extracolsep{\fill}}lccccccc}
\toprule
\textbf{Metric} & \textbf{Adam} & \textbf{AdamW} & \textbf{RAdam} & \textbf{AdamP} & \textbf{Sophia} & \textbf{MARS} & \textbf{AFOR} \\
\midrule
Extra FLOPs \textit{vs.} Adam (\textit{K}) & -- & 0.00 & 0.29 & 15.56 & 17.66 & 10.46 & 19.51 \\
State memory (\textit{KB})       & 13.69 & 13.69 & 13.64 & 13.64 & 13.64 & 20.46 & 14.11 \\
Latency overhead \textit{vs.} Adam (\textit{ms}) & -- & 0.07 & 0.01 & 1.15 & 0.89 & 0.76 & 1.41 \\
\bottomrule
\end{tabular*}
\end{table*}

\subsection{Sensitivity and Complexity Analysis}

\noindent\textbf{Sensitivity analysis.}
AFOR introduces three additional hyperparameters beyond Adam: the lower
bound $\beta_2^{\min}$, the direction weight $w$, and the warm-up length
$T_w$. Their design values ($\beta_2^{\min} = 0.99$, $w = 1.0$,
$T_w = 100$) are fixed before evaluation and are not tuned on the test
splits. Fig.~\ref{fig:sensitivity} shows the mean \textit{Acc} obtained by
varying one hyperparameter at a time on DEAP and ISRUC with EEGNet,
using the same repeated-run protocol as Table~\ref{tab:results}. The design
values achieve the highest observed accuracy on both datasets. Across the tested settings,
\textit{Acc} ranges from 61.51\% to 62.94\% on DEAP and from 66.04\% to
67.57\% on ISRUC. The largest variation is associated with
$\beta_2^{\min}$ on DEAP and with $w$ on ISRUC, while the warm-up
length has a narrower effect on DEAP.

\noindent\textbf{Algorithmic complexity.}
Table~\ref{tab:complexity} reports the measured optimizer-side cost on DEAP with EEGNet, including extra floating-point operations (FLOPs) relative to Adam, the optimizer state memory, and the latency overhead relative to Adam. 
To evaluate this cost, we use identical model weights and reused
gradients, measuring only the warmed-up update phase. AdamW matches Adam in
state size and arithmetic, RAdam adds negligible cost, and MARS uses the most
memory. AFOR increases memory by 3.07\% (14.11 \textit{vs.} 13.69 \textit{KB})
and adds 19.51 \textit{K} FLOPs and 1.41 \textit{ms} per update, while retaining
the same $\Theta(d)$ asymptotic complexity.

For the theoretical orders, Adam applies a constant number of element-wise operations to each parameter tensor. Its moment recursions, initialization correction, and parameter update are as follows:
\begin{equation}
\begin{aligned}
&m_t^{(k)} = \beta_1 m_{t-1}^{(k)} + (1-\beta_1) g_t^{(k)}, \\
&v_t^{(k)} = \beta_2 v_{t-1}^{(k)} + (1-\beta_2) g_t^{(k)} \odot g_t^{(k)}, \\
&\hat m_t^{(k)} = \frac{m_t^{(k)}}{1-\beta_1^t}, \quad
\hat v_t^{(k)} = \frac{v_t^{(k)}}{1-\beta_2^t}, \\
&\theta_t^{(k)} = \theta_{t-1}^{(k)} - \eta_t \frac{\hat m_t^{(k)}}{\sqrt{\hat v_t^{(k)}}+\varepsilon}.
\end{aligned}
\end{equation}
When decoupled weight decay is enabled, it adds one elementwise term of
linear cost and does not change the complexity order.
AFOR keeps this update and adds only a fixed set of tensor-wise terms:
\begin{equation}
\begin{aligned}
\rho_t^{(k)} &=
\left(\frac{\frac{1}{d_k}\sum_{j=1}^{d_k}|m_{t,j}^{(k)}|}
{\max\!\bigl(n_t^{\text{f},(k)}, n_t^{\text{s},(k)}\bigr)+\varepsilon}\right)
 \cdot \bigl(1+w(\bar c_t^{(k)}-1)\bigr), \\
\beta_{2,t}^{(k)} &= \gamma_t\,\beta_{2,t}^{\mathrm{raw},(k)} + (1-\gamma_t)\,\beta_2^{\mathrm{init}}, \\
C_t^{(k)} &= C_{t-1}^{(k)}\beta_{2,t}^{(k)}.
\end{aligned}
\end{equation}

Each term is still linear in tensor size, so both methods remain
$\Theta(d)$. Thus, the measured overhead in Table~\ref{tab:complexity} is a constant-factor cost on top of the same $\Theta(d)$ per-step order as the standard Adam update.

\section{CONVERGENCE AND STABILITY ANALYSIS}
The analysis below establishes a conditional finite-time stability and stationarity bound. 
We analyze the concatenated parameter vector, with the per-tensor
preconditioners assembled into the block-diagonal matrix $H_t$. Let
$f(\theta)=\mathbb{E}_{\xi}[\ell(\theta;\xi)]$ be lower bounded and
$L$-smooth, where $\xi$ denotes the sampled data and
$f_\star:=\inf_\theta f(\theta)>-\infty$. For the filtration
$\mathcal{F}_{t-1}$ generated before drawing $g_t$, assume
\begin{equation}
\begin{aligned}
&\mathbb{E}[g_t\mid\mathcal{F}_{t-1}]
=\nabla f(\theta_{t-1}),\\
&\mathbb{E}\!\left[
\|g_t-\nabla f(\theta_{t-1})\|_2^2
\mid\mathcal{F}_{t-1}\right] \leq\sigma_{g,t}^2,
\end{aligned}
\label{eq:afor_gradient}
\end{equation}
with a uniform bound $\|g_t\|_\infty\leq G$ almost surely. For every tensor,
$0<\beta_2^{\min}\leq\beta_{2,t}\leq\beta_2^{\max}<1$.
In the implementation, $\beta_2^{\min}=0.99$ and
$\beta_2^{\max}=\beta_2^{\text{init}}=0.999$; thus the gate and sigmoid
mapping in Eqs.~\eqref{eq:beta2_raw}--\eqref{eq:gate} satisfy this bound.
Consequently, $0<C_t\leq(\beta_2^{\max})^t$ and
$1-C_t\geq1-(\beta_2^{\max})^t>0$, so the corrected denominator is well
defined.

\noindent\textbf{Tracking condition for the AFOR direction.}
For the AFOR preconditioned direction $p_t=H_t\hat m_t$ with
$H_t=\operatorname{Diag}((\sqrt{\hat v_t}+\varepsilon)^{-1})$, let
$u_t=\nabla f(\theta_{t-1})$ and $\nu_t^g=g_t-u_t$.
The bias-corrected first moment has deterministic weights
\begin{equation}
\hat m_t=\sum_{i=1}^{t}\omega_{t,i}g_i,\qquad
\omega_{t,i}:=\frac{(1-\beta_1)\beta_1^{\,t-i}}{1-\beta_1^{\,t}},
\label{eq:afor_momentum_weights}
\end{equation}
where $\sum_{i=1}^{t}\omega_{t,i}=1$.
Assume
\begin{equation}
\mathbb{E}\!\left[
\left\|\sum_{i=1}^{t}\omega_{t,i}(u_i-u_t)\right\|_2^2
\right]\leq\Delta_t^2 .
\label{eq:afor_gradient_drift}
\end{equation}
Using the martingale-difference property of $\nu_i^g$ and
$\|x+y\|_2^2\leq2\|x\|_2^2+2\|y\|_2^2$ gives
\begin{equation}
\mathbb{E}\!\left[\|\hat m_t-u_t\|_2^2\right]\leq\kappa_t^2,\quad
\kappa_t^2:=2\Delta_t^2+
2\sum_{i=1}^{t}\omega_{t,i}^2\sigma_{g,i}^2 .
\label{eq:afor_tracking}
\end{equation}
Let $\bar\kappa^2:=\sup_t\kappa_t^2<\infty$. The asymptotic result below
also requires $\sum_t\eta_t\kappa_t^2<\infty$, which bounded noise alone
does not guarantee. Here $u_i$ is the gradient on the subjects sampled at step
$i$, so $\Delta_t^2$ measures subject shift not random noise. The
tensor-wise residual $e_t$ of Eq.~\eqref{eq:noise_inst} is the empirical
counterpart of the tracking error bounded by $\kappa_t^2$, so RASS responds to
the quantity that governs the descent bound in
Eq.~\eqref{eq:afor_derived_bounds}.

\noindent\textbf{Lemma 1 (Conditional finite-time weighted stationarity under tracking control).}
Let $v_0=0$, $v_t=\beta_{2,t}v_{t-1}
+(1-\beta_{2,t})g_t\odot g_t$, and
$C_t=\prod_{s=1}^{t}\beta_{2,s}$. Unrolling the recursion gives
\begin{equation}
\begin{aligned}
\hat v_t&=\frac{v_t}{1-C_t}
=\sum_{i=1}^{t}\pi_{t,i}\,(g_i\odot g_i),\\
\pi_{t,i}&:=
\frac{(1-\beta_{2,i})\prod_{j=i+1}^{t}\beta_{2,j}}{1-C_t},
\qquad\sum_{i=1}^{t}\pi_{t,i}=1,
\end{aligned}
\label{eq:afor_convex_form}
\end{equation}
where the sum follows by telescoping and $\pi_{t,i}\geq0$. Hence
$0\leq\hat v_t\leq G^2$ elementwise. The same argument gives
$|\hat m_t|\leq G$ and
$\|\hat m_t/(\sqrt{\hat v_t}+\varepsilon)\|_\infty\leq G/\varepsilon$.
Because $\beta_{2,t}$ is computed from the current gradient, each
$\pi_{t,i}$ may depend on the observations it weights, both through
$\beta_{2,i}$ and through later adaptive coefficients. Thus,
Eq.~\eqref{eq:afor_convex_form} is an algebraic convex decomposition rather
than an unbiased second-moment estimator. The tracking condition is therefore
imposed directly on the first-moment direction and is not inferred from
Lemma~1.

Let $\tau_t=\hat m_t-u_t$. Lemma~1 yields
$\alpha I\preceq H_t\preceq M I$, where $I$ is the identity matrix,
$\preceq$ denotes the Loewner order, and
$\alpha:=1/(G+\varepsilon)$, $M:=1/\varepsilon$.
Combining this bound with Eq.~\eqref{eq:afor_tracking} gives
\begin{equation}
\begin{aligned}
\mathbb{E}[\langle u_t,p_t\rangle]
&\geq \frac{\alpha}{2}\mathbb{E}[\|u_t\|_2^2]
-\frac{M^2}{2\alpha}\kappa_t^2,\\
\mathbb{E}[\|p_t\|_2^2]
&\leq 2M^2\left(\mathbb{E}[\|u_t\|_2^2]+\kappa_t^2\right).
\end{aligned}
\label{eq:afor_derived_bounds}
\end{equation}
Set $a_0:=\alpha/2$, $r_t^0:=M^2\kappa_t^2/(2\alpha)$, and
$b_0:=2M^2(1+\bar\kappa^2)$. Here $\kappa_t^2$ and $r_t^0$ are
deterministic bounds, not realized tracking errors.

The decoupled weight-decay update is
\begin{equation}
\theta_t=(1-\eta_t\lambda)\theta_{t-1}-\eta_t p_t
=\theta_{t-1}-\eta_t q_t,
\label{eq:afor_decay_update}
\end{equation}
with $q_t:=p_t+\lambda\theta_{t-1}$.
Let $N_\theta$ be the number of scalar parameters. Since
$\|p_t\|_2\leq D:=\sqrt{N_\theta}G/\varepsilon$, for $\lambda>0$ and
$0\leq\eta_t\lambda\leq1$ we have
$\|\theta_t\|_2\leq(1-\eta_t\lambda)\|\theta_{t-1}\|_2+\eta_tD$, hence
$\|\theta_t\|_2\leq R:=\max(\|\theta_0\|_2,D/\lambda)$.

Because decoupled decay is not a preconditioned gradient estimator of
$f(\theta)+\lambda\|\theta\|_2^2/2$, the result below is stated for $f$.
Define $\chi_\lambda:=0$ for $\lambda=0$ and $\chi_\lambda:=\lambda R$ for
$\lambda>0$.

\noindent\textbf{Theorem 1 (Conditional weighted stationarity).}
Let $a_\lambda:=a_0/2$, $b_\lambda:=2b_0+2\chi_\lambda^2$, and
$r_t^\lambda:=r_t^0+\chi_\lambda^2/(2a_0)$. If
$0<\eta_t\leq a_\lambda/(Lb_\lambda)$ and $\eta_t\lambda\leq1$, then
\begin{equation}
\begin{aligned}
\frac{\sum_{t=1}^{N}\eta_t\,
\mathbb{E}[\|\nabla f(\theta_{t-1})\|_2^2]}
{\sum_{t=1}^{N}\eta_t}
&\leq{}
\frac{2(f(\theta_0)-f_\star)}
{a_\lambda\sum_{t=1}^{N}\eta_t}\\
&+\frac{\sum_{t=1}^{N}(2\eta_t r_t^\lambda+Lb_\lambda\eta_t^2)}
{a_\lambda\sum_{t=1}^{N}\eta_t}.
\end{aligned}
\label{eq:afor_finite_bound}
\end{equation}

\noindent\textit{Proof.}
The preceding bounds and $\lambda\|\theta_{t-1}\|_2\leq\chi_\lambda$ imply
\begin{equation}
\begin{aligned}
\mathbb{E}[\langle u_t,q_t\rangle]
&\geq a_0\mathbb{E}[\|u_t\|_2^2]
-r_t^0-\chi_\lambda\mathbb{E}[\|u_t\|_2]\\
&\geq a_\lambda\mathbb{E}[\|u_t\|_2^2]-r_t^\lambda,\\
\mathbb{E}[\|q_t\|_2^2]
&\leq b_\lambda\left(1+\mathbb{E}[\|u_t\|_2^2]\right).
\end{aligned}
\label{eq:afor_decay_alignment_bound}
\end{equation}
The $L$-smoothness inequality and $\theta_t=\theta_{t-1}-\eta_tq_t$ give
\begin{equation}
\begin{aligned}
\mathbb{E}[f(\theta_t)]
\leq{}&\mathbb{E}[f(\theta_{t-1})]
-\left(a_\lambda\eta_t-\frac{Lb_\lambda\eta_t^2}{2}\right)
\mathbb{E}[\|u_t\|_2^2]\\
&+\eta_t r_t^\lambda+\frac{Lb_\lambda\eta_t^2}{2}.
\end{aligned}
\label{eq:afor_descent}
\end{equation}
Since $\eta_t\leq a_\lambda/(Lb_\lambda)$, summing this inequality,
using $f(\theta_N)\geq f_\star$, and dividing by
$\sum_{t=1}^{N}\eta_t$ proves Eq.~\eqref{eq:afor_finite_bound}. This is the
finite-time result: it allows nonzero tracking residuals and yields a
neighborhood bound rather than vanishing stationarity.

The asymptotic zero-stationarity statement requires, in addition,
$\sum_t\eta_t=\infty$ and $\sum_t\eta_t^2<\infty$, together with
$\sum_t\eta_t r_t^0<\infty$. The last condition is not implied by bounded
noise: if $\sigma_{g,t}^2\geq\sigma^2>0$ and $\beta_1\in(0,1)$, then
$\sum_i\omega_{t,i}^2\to(1-\beta_1)/(1+\beta_1)>0$, so $r_t^0$ stays
bounded away from zero and $\sum_t\eta_t r_t^0<\infty$ would force
$\sum_t\eta_t<\infty$. The asymptotic results below therefore hold in the
vanishing-noise, vanishing-drift regime and do not extend to persistent
noise. When $\lambda=0$,
\begin{equation}
\frac{\sum_{t=1}^{N}\eta_t\,
\mathbb{E}[\|\nabla f(\theta_{t-1})\|_2^2]}
{\sum_{t=1}^{N}\eta_t}
\longrightarrow0\qquad(N\rightarrow\infty).
\label{eq:afor_stationarity}
\end{equation}
Thus $\liminf_{t\rightarrow\infty}
\mathbb{E}[\|\nabla f(\theta_{t-1})\|_2^2]=0$. For $\lambda>0$, the
corresponding bound is
\begin{equation}
\limsup_{N\rightarrow\infty}
\frac{\sum_{t=1}^{N}\eta_t\,
\mathbb{E}[\|\nabla f(\theta_{t-1})\|_2^2]}
{\sum_{t=1}^{N}\eta_t}
\leq\frac{2\chi_\lambda^2}{a_0^2}
=\frac{2\lambda^2R^2}{a_0^2}.
\label{eq:afor_weight_decay_gap}
\end{equation}
Therefore, nonzero decoupled weight decay produces an explicit stationarity
gap.

\noindent\textbf{Fixed learning rate.}
If $\eta_t=\eta\leq a_\lambda/(Lb_\lambda)$ and
$r_t^0\leq\bar r^0$, then
\begin{equation}
\begin{aligned}
\frac{1}{N}\sum_{t=1}^{N}
\mathbb{E}[\|\nabla f(\theta_{t-1})\|_2^2]
&\leq{}
\frac{2(f(\theta_0)-f_\star)}{a_\lambda\eta N}\\
&+\frac{2}{a_\lambda}
\left(\bar r^0+\frac{\chi_\lambda^2}{2a_0}\right)
+\frac{Lb_\lambda}{a_\lambda}\eta .
\end{aligned}
\label{eq:afor_fixed_stepsize}
\end{equation}
For the experimental $\lambda=10^{-4}$, this is a finite-step neighborhood
guarantee for the original objective, conditional on the tracking requirement
of Eq.~\eqref{eq:afor_tracking}.

\section{CONCLUSION}

Gradient statistics in cross-subject EEG decoding shift across layers, subjects, and training phases, making the task inherently nonstationary. Standard optimizers nevertheless apply one fixed second-moment decay coefficient to every tensor. AFOR removes this assumption by treating the coefficient as a tensor-wise, time-varying state estimated online from local gradient quality. 

Under a strict cross-subject protocol, AFOR improves over Adam across three
benchmarks spanning different domains and four representative backbones,
achieving the best average performance among all compared optimizers, with
ablations confirming that each adaptive component contributes and that the
added computational cost is only a constant factor.

While AFOR addresses the fixed second-moment memory limitation, two aspects remain open for future investigation. First, the guarantee in Section~VI is conditional on the gradient-drift requirement and rests on worst-case constants, so it is qualitative rather than a quantitative
certificate for the reported runs. Second, the three additional hyperparameters are fixed to defaults validated by our sensitivity study, and adjusting them automatically would reduce manual tuning. We further plan to extend the same memory regulation to clinical applications such as brain disease prediction and EEG-based multimodal analysis.

\bibliographystyle{IEEEtran}
\bibliography{reference}

\begin{thebibliography}{10}
\providecommand{\url}[1]{#1}
\csname url@samestyle\endcsname
\providecommand{\newblock}{\relax}
\providecommand{\bibinfo}[2]{#2}
\providecommand{\BIBentrySTDinterwordspacing}{\spaceskip=0pt\relax}
\providecommand{\BIBentryALTinterwordstretchfactor}{4}
\providecommand{\BIBentryALTinterwordspacing}{\spaceskip=\fontdimen2\font plus
\BIBentryALTinterwordstretchfactor\fontdimen3\font minus \fontdimen4\font\relax}
\providecommand{\BIBforeignlanguage}[2]{{%
\expandafter\ifx\csname l@#1\endcsname\relax
\typeout{** WARNING: IEEEtran.bst: No hyphenation pattern has been}%
\typeout{** loaded for the language `#1'. Using the pattern for}%
\typeout{** the default language instead.}%
\else
\language=\csname l@#1\endcsname
\fi
#2}}
\providecommand{\BIBdecl}{\relax}
\BIBdecl

\bibitem{marino2026human}
M.~Marino and D.~Mantini, ``Human brain imaging with high-density electroencephalography: Techniques and applications,'' \emph{The Journal of Physiology}, vol. 604, no.~2, pp. 783--812, 2026.

\bibitem{eegnet}
V.~J. Lawhern, A.~J. Solon, N.~R. Waytowich, S.~M. Gordon, C.~P. Hung, and B.~J. Lance, ``Eegnet: A compact convolutional neural network for eeg-based brain--computer interfaces,'' \emph{Journal of Neural Engineering}, vol.~15, no.~5, p. 056013, 2018.

\bibitem{2022TSception}
Y.~Ding, N.~Robinson, S.~Zhang, Q.~Zeng, and C.~Guan, ``Tsception: Capturing temporal dynamics and spatial asymmetry from eeg for emotion recognition,'' \emph{IEEE Transactions on affective computing}, vol.~14, no.~3, pp. 2238--2250, 2022.

\bibitem{acrnn}
W.~Tao, C.~Li, R.~Song, J.~Cheng, Y.~Liu, F.~Wan, and X.~Chen, ``Eeg-based emotion recognition via channel-wise attention and self attention,'' \emph{IEEE Transactions on Affective Computing}, vol.~14, no.~4, pp. 3074--3087, 2023.

\bibitem{2023EEGRNN2}
A.~Sam, R.~Boostani, S.~Hashempour, M.~Taghavi, and S.~Sanei, ``Depression identification using eeg signals via a hybrid of lstm and spiking neural networks,'' \emph{IEEE Transactions on Neural Systems and Rehabilitation Engineering}, vol.~31, pp. 4725--4737, 2023.

\bibitem{nsaldgat}
Y.~Yang, Z.~Wang, Y.~Song, Z.~Jia, B.~Wang, T.-P. Jung, and F.~Wan, ``Exploiting the intrinsic neighborhood semantic structure for domain adaptation in eeg-based emotion recognition,'' \emph{IEEE Transactions on Affective Computing}, vol.~16, no.~3, pp. 2466--2478, 2025.

\bibitem{EEGGNN2}
C.-M. Cui, H.-Y. Chen, M.-S. Chen, J.~Li, Z.~Tong, C.~Fang, C.-D. Wang, and Y.~Cai, ``“pre-train, prompt” framework to boost graph neural networks performance in eeg analysis,'' \emph{IEEE Journal of Biomedical and Health Informatics}, vol.~29, no.~7, 2025.

\bibitem{hslt}
Z.~Wang, Y.~Wang, C.~Hu, Z.~Yin, and Y.~Song, ``Transformers for eeg-based emotion recognition: A hierarchical spatial information learning model,'' \emph{IEEE Sensors Journal}, vol.~22, no.~5, pp. 4359--4368, 2022.

\bibitem{eegtrans2}
C.-H. Chuang, K.-Y. Chang, C.-S. Huang, and A.-M. Bessas, ``Augmenting brain-computer interfaces with art: An artifact removal transformer for reconstructing multichannel eeg signals,'' \emph{NeuroImage}, vol. 310, p. 121123, 2025.

\bibitem{EEGBCI2025}
Y.~Ding, Y.~Li, H.~Sun, R.~Liu, C.~Tong, C.~Liu, X.~Zhou, and C.~Guan, ``Eeg-deformer: A dense convolutional transformer for brain-computer interfaces,'' \emph{IEEE Journal of Biomedical and Health Informatics}, vol.~29, no.~3, pp. 1909--1918, 2025.

\bibitem{yang2026dual}
Y.~Yang, W.~Wang, K.~Shi, Y.~Xie, N.~Zhou, S.~Wen, M.~Zhu, and B.~Chen, ``A dual-stream regional feature learning and adaptive fusion method for electroencephalogram-based emotion recognition,'' \emph{Engineering Applications of Artificial Intelligence}, vol. 164, p. 113250, 2026.

\bibitem{yang2026prototypical}
Y.~Yang, C.~Sun, R.~Lyu, J.~Wang, Z.~Wang, X.~Chen, C.-T. Lin, T.-P. Jung, and F.~Wan, ``Prototypical contrastive learning with temporal dynamic graph convolutional network for eeg-based emotion recognition,'' \emph{IEEE Transactions on Affective Computing}, 2026.

\bibitem{li2026cross}
T.~Li, Y.~Yan, F.~Dou, W.~Song, and X.~Zhang, ``Cross-subject generalization for eeg decoding: a survey of deep learning methods,'' \emph{Progress in Biomedical Engineering}, vol.~8, no.~2, p. 022013, 2026.

\bibitem{ng2024subject}
H.~W. Ng and C.~Guan, ``Subject-independent meta-learning framework towards optimal training of eeg-based classifiers,'' \emph{Neural Networks}, vol. 172, p. 106108, 2024.

\bibitem{libeer}
H.~Liu, S.~Yang, Y.~Zhang, M.~Wang, F.~Gong, C.~Xie, G.~Liu, Z.~Liu, Y.-J. Liu, B.-L. Lu, and D.~Zhang, ``Libeer: A comprehensive benchmark and algorithm library for eeg-based emotion recognition,'' \emph{IEEE Transactions on Affective Computing}, vol.~16, no.~4, pp. 3596--3613, 2025.

\bibitem{adam}
D.~P. Kingma and J.~Ba, ``Adam: A method for stochastic optimization,'' in \emph{International Conference on Learning Representations (ICLR)}, 2015.

\bibitem{adamw}
I.~Loshchilov and F.~Hutter, ``Decoupled weight decay regularization,'' in \emph{International Conference on Learning Representations (ICLR)}, 2019.

\bibitem{zhu2026phy}
H.~Zhu, L.~Chen, Y.~Fu, M.~A. El-Yacoubi, and M.~Shang, ``Physiological prior-driven label enhancement for cross-subject eeg emotion recognition,'' \emph{arXiv preprint arXiv:2607.15566}, 2026.

\bibitem{radam}
L.~Liu, H.~Jiang, P.~He, W.~Chen, X.~Liu, J.~Gao, and J.~Han, ``On the variance of the adaptive learning rate and beyond,'' in \emph{International Conference on Learning Representations (ICLR)}, 2020.

\bibitem{adamp}
B.~Heo, S.~Chun, S.~J. Oh, D.~Han, S.~Yun, G.~Kim, Y.~Uh, and J.-W. Ha, ``Adamp: Slowing down the slowdown for momentum optimizers on scale-invariant weights,'' in \emph{International Conference on Learning Representations (ICLR)}, 2021.

\bibitem{sophia}
H.~Liu, Z.~Li, D.~Hall, P.~Liang, and T.~Ma, ``Sophia: A scalable stochastic second-order optimizer for language model pre-training,'' in \emph{International Conference on Learning Representations (ICLR)}, 2024.

\bibitem{mars}
H.~Yuan, Y.~Liu, S.~Wu, X.~Zhou, and Q.~Gu, ``Mars: unleashing the power of variance reduction for training large models,'' in \emph{International Conference on Machine Learning (ICML)}, 2025.

\bibitem{paleologu2008robust}
C.~Paleologu, J.~Benesty, and S.~Ciochin{\u{a}}, ``A robust variable forgetting factor recursive least-squares algorithm for system identification,'' \emph{IEEE Signal Processing Letters}, vol.~15, pp. 597--600, 2008.

\bibitem{huang2019nostalgic}
H.~Huang, C.~Wang, and B.~Dong, ``Nostalgic adam: Weighting more of the past gradients when designing the adaptive learning rate,'' in \emph{Proceedings of the Twenty-Eighth International Joint Conference on Artificial Intelligence (IJCAI)}.\hskip 1em plus 0.5em minus 0.4em\relax International Joint Conferences on Artificial Intelligence Organization, 2019.

\bibitem{zhu2026bimoe}
H.~Zhu, L.~Chen, and M.~Shang, ``Bimoe: Brain-inspired experts for eeg-dominant affective state recognition,'' \emph{arXiv preprint arXiv:2603.29205}, 2026.

\bibitem{ding2025emt}
Y.~Ding, C.~Tong, S.~Zhang, M.~Jiang, Y.~Li, K.~J. Lim, and C.~Guan, ``Emt: A novel transformer for generalized cross-subject eeg emotion recognition,'' \emph{IEEE Transactions on Neural Networks and Learning Systems}, vol.~36, no.~6, pp. 10\,381--10\,393, 2025.

\bibitem{phang2025explainable}
C.-R. Phang and A.~Hirata, ``Explainable multiscale temporal convolutional neural network model for sleep stage detection based on electroencephalogram activities,'' \emph{Journal of Neural Engineering}, vol.~22, no.~2, p. 026010, 2025.

\bibitem{wang2026brastorm}
Y.~Wang, D.-H. Lee, and X.~Bruce, ``Brastorm: A dual-branch self-supervised framework for eeg representation learning via input-level spatio-temporal decomposition,'' in \emph{Proceedings of the AAAI Conference on Artificial Intelligence}, vol.~40, no.~21, 2026.

\bibitem{saibene2024mireview}
A.~Saibene, H.~Ghaemi, and E.~Dagdevir, ``Deep learning in motor imagery eeg signal decoding: A systematic review,'' \emph{Neurocomputing}, vol. 610, p. 128577, 2024.

\bibitem{rao2024wearable}
Z.~Rao, J.~Zhu, Z.~Lu, R.~Zhang, K.~Li, Z.~Guan, and Y.~Li, ``A wearable brain-computer interface with fewer eeg channels for online motor imagery detection,'' \emph{IEEE Transactions on Neural Systems and Rehabilitation Engineering}, vol.~32, pp. 4143--4154, 2024.

\bibitem{kostulin2026eeg}
D.~Kostulin, P.~Shaposhnikov, A.~K. Ekizyan, I.~Shevchenko, D.~Shaposhnikov, I.~Shcherban, and V.~Kiroy, ``Eeg-based brain-computer interface (bci) dataset for directional word recognition,'' \emph{Scientific Data}, vol.~13, no.~1, p. 1195, 2026.

\bibitem{rakhmatulin2024cnn}
I.~Rakhmatulin, M.-S. Dao, A.~Nassibi, and D.~Mandic, ``Exploring convolutional neural network architectures for eeg feature extraction,'' \emph{Sensors}, vol.~24, no.~3, p. 877, 2024.

\bibitem{eegnetmsd2023}
R.~Fu, Z.~Wang, S.~Wang, X.~Xu, J.~Chen, and G.~Wen, ``Eegnet-msd: A sparse convolutional neural network for efficient eeg-based intent decoding,'' \emph{IEEE Sensors Journal}, vol.~23, no.~17, pp. 19\,684--19\,691, 2023.

\bibitem{hajisafi2024neurognn}
A.~Hajisafi, H.~Lin, Y.-Y. Chiang, and C.~Shahabi, ``Dynamic gnns for precise seizure detection and classification from eeg data,'' in \emph{Advances in Knowledge Discovery and Data Mining}, ser. Lecture Notes in Computer Science, vol. 14648.\hskip 1em plus 0.5em minus 0.4em\relax Springer, 2024, pp. 207--220.

\bibitem{2011deap}
S.~Koelstra, C.~Muhl, M.~Soleymani, J.-S. Lee, A.~Yazdani, T.~Ebrahimi, T.~Pun, A.~Nijholt, and I.~Patras, ``Deap: A database for emotion analysis using physiological signals,'' \emph{IEEE Transactions on Affective Computing}, vol.~3, no.~1, pp. 18--31, 2011.

\bibitem{seed}
W.-L. Zheng and B.-L. Lu, ``Investigating critical frequency bands and channels for eeg-based emotion recognition with deep neural networks,'' \emph{IEEE Transactions on Autonomous Mental Development}, vol.~7, no.~3, pp. 162--175, 2015.

\bibitem{bci42a}
M.~Tangermann, K.-R. Mueller, A.~Aertsen, N.~Birbaumer, C.~Braun, C.~Brunner, R.~Leeb, C.~Mehring, K.~J. Miller, G.~R. Mueller-Putz, G.~Nolte, G.~Pfurtscheller, H.~Preissl, G.~Schalk, A.~Schlogl, C.~Vidaurre, S.~Waldert, and B.~Blankertz, ``Review of the bci competition iv,'' \emph{Frontiers in Neuroscience}, vol.~6, p.~55, 2012.

\bibitem{isruc}
S.~Khalighi, T.~Sousa, J.~M. Santos, and U.~Nunes, ``Isruc-sleep: A comprehensive public dataset for sleep researchers,'' \emph{Computer Methods and Programs in Biomedicine}, vol. 124, pp. 180--192, 2016.

\bibitem{Holm1979ASS}
S.~Holm, ``A simple sequentially rejective multiple test procedure,'' \emph{Scandinavian Journal of Statistics}, vol.~6, pp. 65--70, 1979.

\bibitem{li2025mixln}
P.~Li, L.~Yin, and S.~Liu, ``Mix-ln: Unleashing the power of deeper layers by combining pre-ln and post-ln,'' in \emph{International Conference on Learning Representations (ICLR)}, 2025.

\bibitem{gray2024normalization}
G.~Gray, A.~Tiwari, S.~Bergsma, and J.~Hestness, ``Normalization layer per-example gradients are sufficient to predict gradient noise scale in transformers,'' in \emph{Advances in Neural Information Processing Systems (NeurIPS)}, 2024.

\end{thebibliography}

\end{document}